\documentclass[letterpaper]{article} % DO NOT CHANGE THIS
\usepackage{aaai2026}  % DO NOT CHANGE THIS
\usepackage{times}  % DO NOT CHANGE THIS
\usepackage{helvet}  % DO NOT CHANGE THIS
\usepackage{courier}  % DO NOT CHANGE THIS
\usepackage[hyphens]{url}  % DO NOT CHANGE THIS
\usepackage{graphicx} % DO NOT CHANGE THIS
\usepackage{natbib}  % DO NOT CHANGE THIS AND DO NOT ADD ANY OPTIONS TO IT
\usepackage{caption} % DO NOT CHANGE THIS AND DO NOT ADD ANY OPTIONS TO IT
\usepackage{algorithm}
\usepackage{algorithmic}
\usepackage{amsmath}
\usepackage{multirow}
\usepackage{makecell}
\usepackage{newfloat}
\usepackage{listings}
\DeclareCaptionStyle{ruled}{labelfont=normalfont,labelsep=colon,strut=off} % DO NOT CHANGE THIS
\floatstyle{ruled}
\newfloat{listing}{tb}{lst}{}
\floatname{listing}{Listing}
\title{Exploring Reliable Spatiotemporal Dependencies for Efficient Visual Tracking}
\author{
	Junze Shi\textsuperscript{\rm 1,2,3}, Yang Yu\textsuperscript{\rm 1,2}, Jian Shi\textsuperscript{\rm 1,2,3}, Haibo Luo\textsuperscript{\rm 1,2}\thanks{Corresponding author.}
}
\affiliations{
	\textsuperscript{\rm 1}Key Laboratory of Opto-Electronic Information Processing, Chinese Academy of Sciences\\
	\textsuperscript{\rm 2}Shenyang Institute of Automation, Chinese Academy of Sciences\\
	\textsuperscript{\rm 3}University of Chinese Academy of Sciences
	shijunze@sia.cn, yuyang@sia.cn, shijian@sia.cn, luohb@sia.cn
}

\usepackage{bibentry}
\begin{document}

\maketitle

\begin{abstract}
Recent advances in transformer-based lightweight object tracking have established new standards across benchmarks, leveraging the global receptive field and powerful feature extraction capabilities of attention mechanisms. Despite these achievements, existing methods universally employ sparse sampling during training—utilizing only one template and one search image per sequence—which fails to comprehensively explore spatiotemporal information in videos. This limitation constrains performance and causes the gap between lightweight and high-performance trackers. To bridge this divide while maintaining real-time efficiency, we propose STDTrack, a framework that pioneers the integration of reliable spatiotemporal dependencies into lightweight trackers. Our approach implements dense video sampling to maximize spatiotemporal information utilization. We introduce a temporally propagating spatiotemporal token to guide per-frame feature extraction. To ensure comprehensive target state representation, we design the Multi-frame Information Fusion Module (MFIFM), which augments current dependencies using historical context. The MFIFM operates on features stored in our constructed Spatiotemporal Token Maintainer (STM), where a quality-based update mechanism ensures information reliability. Considering the scale variation among tracking targets, we develop a multi-scale prediction head to dynamically adapt to objects of different sizes. Extensive experiments demonstrate state-of-the-art results across six benchmarks. Notably, on GOT-10k, STDTrack rivals certain high-performance non-real-time trackers ($e.g.$, MixFormer) while operating at 192 FPS (GPU) and 41 FPS (CPU).
\end{abstract}

% Uncomment the following to link to your code, datasets, an extended version or similar.
% You must keep this block between (not within) the abstract and the main body of the paper.
% \begin{links}
%     \link{Code}{https://aaai.org/example/code}
%     \link{Datasets}{https://aaai.org/example/datasets}
%     \link{Extended version}{https://aaai.org/example/extended-version}
% \end{links}
\section{Introduction}
Visual object tracking, a cornerstone task in computer vision, aims to continuously localize arbitrary objects in video sequences based on their initial states. This technology finds widespread applications in autonomous driving systems, pedestrian detection, and unmanned aerial vehicle (UAV) operations. However, prior research has predominantly focused on enhancing tracker accuracy at the expense of operational efficiency~\cite{ARTrack,SeqTrack,ROMTrack,EVPTrack,LMTrack}, rendering most high-performance trackers impractical for deployment in resource-constrained environments. Although numerous lightweight tracking models and methodologies have been proposed~\cite{HCAT,E.T.Track,SiamABC}, existing efficient trackers typically exhibit significant performance degradation, resulting in a substantial performance gap compared to mainstream non-real-time counterparts.

Early researchers attempt to develop real-time trackers predominantly focused on CNN-based architectures~\cite{lighttrack,FEAR}, achieving notable computational efficiency. However, these approaches suffer from insufficient interaction between template and search regions, often leading to suboptimal tracking accuracy. To address this limitation, transformer has been introduced into lightweight trackers to enhance model performance through template-guided feature extraction~\cite{FERMT,SMAT}. However, there is still a significant performance gap compared to state-of-the-art non-real-time trackers. In this work, we aim to bridge the performance gap between lightweight and advanced non-real-time trackers while maintaining real-time capability.

\begin{figure}[t]
	\centering
	\includegraphics[width=0.9\columnwidth]{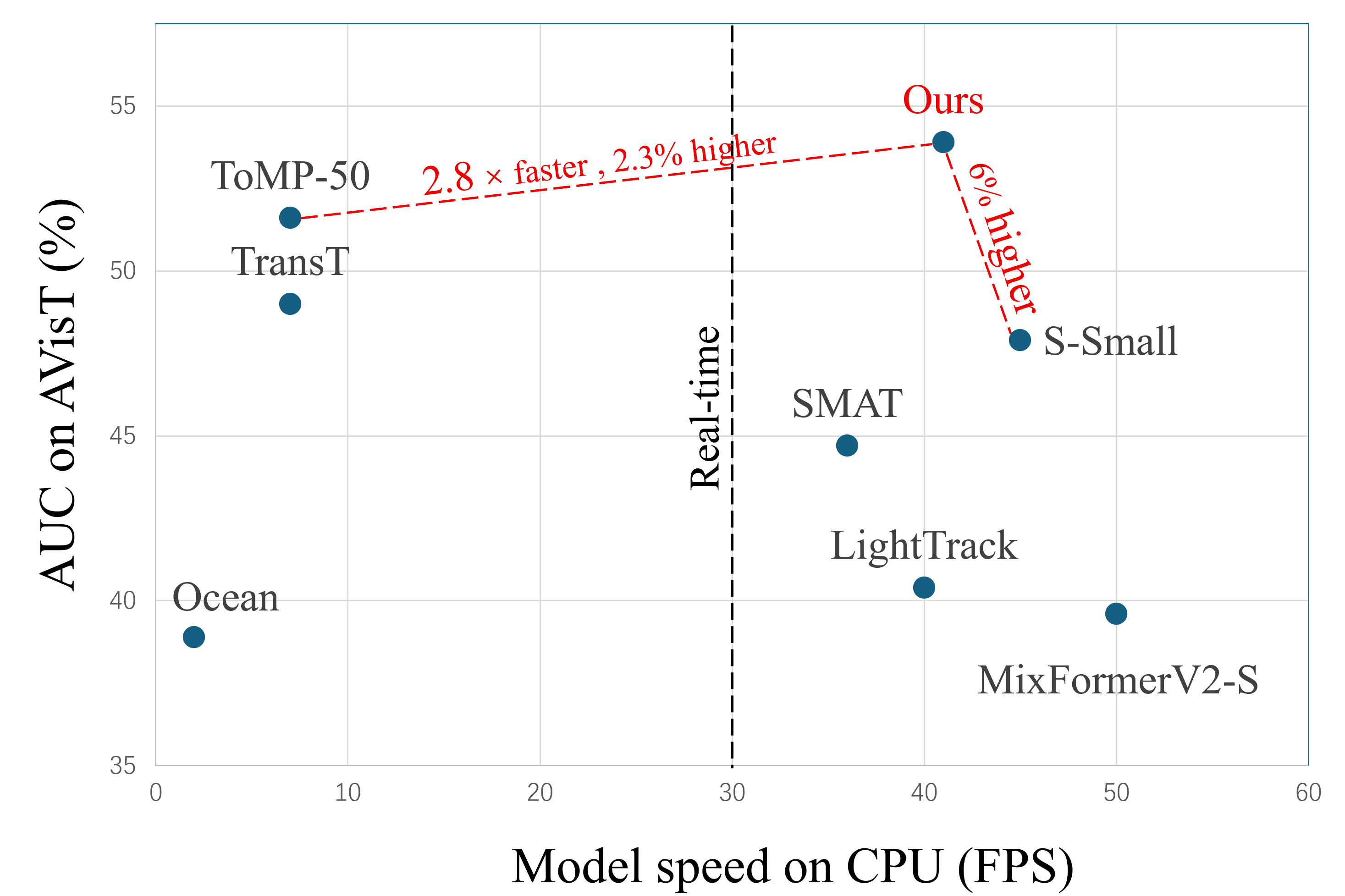} % Reduce the figure size so that it is slightly narrower than the column. Don't use precise values for figure width.This setup will avoid overfull boxes.
	\caption{Comparison of our STDTrack with others on the AVisT dataset~\cite{AVisT} on a CPU. The success score (AUC) (vertical axis) and speed (horizontal axis) are shown. Our tracker achieves substantial accuracy improvement over other state-of-the-art efficient trackers. }
	\label{auc_vs_speed}
\end{figure}

We observe that existing lightweight trackers primarily improve performance by enhancing feature extraction capabilities ($e.g.$, FERMT~\cite{FERMT} boosts feature representation through optimized relational modeling in attention mechanisms). These methods typically sample only one template and one search image per sequence during training, resulting in underutilization of spatiotemporal features through sparse sampling. To address this issue, we propose STDTrack - an innovative lightweight tracking framework incorporating reliable spatiotemporal dependencies. Specifically, our method implements dense sampling of video sequences during training to enhance dataset utilization. For each sampled frame, we introduce a spatiotemporal token to encapsulate target-specific information. To fully exploit spatiotemporal correlations and enable historical state guidance for current frame tracking, we propose a Multi-frame Information Fusion Module (MFIFM).

Furthermore, to enhance the reliability of spatiotemporal dependencies, we construct a Spatiotemporal Token Maintainer (STM) that dynamically assesses the quality of all tokens and prunes the lowest-quality ones upon the introduction of new tokens. This mechanism ensures persistent guidance from historically reliable dependencies during tracking, significantly improving operational stability and robustness. Additionally, we devise a multi-scale prediction head to enhance target size adaptability of the tracker, while employing structural re-parameterization techniques~\cite{RepVGG,RepVGG_modify} during inference to mitigate computational overhead caused by increased parameters.

Based on the above work, our approach achieves substantial performance improvements while maintaining real-time capability. As illustrated in Figure \ref{auc_vs_speed}, our proposed STDTrack attains an AUC of $53.9\%$ on the AVisT benchmark, surpassing the recent state-of-the-art tracker S-Small~\cite{SiamABC} by $6\%$. Meanwhile, STDTrack outperforms the high-performance tracker ToMP-50~\cite{ToMP} while running $2.8 \times$ faster.

Our main contributions are summarized as follows:
\begin{itemize}
	\item We propose a novel lightweight tracking framework STDTrack that integrating reliable spatiotemporal dependencies during tracking. Compared to related work, for the first time, we incorporate continuous spatiotemporal information into lightweight tracking model.
	\item To obtain reliable spatiotemporal dependencies, we design a Multi-frame Information Fusion Module (MFIFM) to exploit contextual information across video sequences. Concurrently, a Spatiotemporal Token Maintainer (STM) is constructed to enhance the reliability of propagated dependencies.
	\item A multi-scale prediction head is proposed to improve the tracker's adaptability to targets of varying sizes.
	\item Extensive experiments demonstrate that our tracker achieves outstanding performance, attaining SOTA results among real-time trackers across multiple benchmarks. Notably, our STDTrack even surpasses the high-performance tracker MixFormer on the GOT-10k dataset.
\end{itemize}

\section{Related Work}
\textbf{Visual Object Tracking based on Transformer.} In the early stage of visual tracking research, practitioners predominantly employed shared-parameter Siamese networks~\cite{SiamFC,SiamRPN} to extract target and search region features respectively, followed by cross-correlation operations for target localization. However, these methods inherently suffer from insufficient template-search interaction, fundamentally limiting their potential for high-performance tracking. Recent advancements in transformer architectures~\cite{Vit,SwinTransformer} have catalyzed a paradigm shift, with increasing research efforts dedicated to constructing transformer-based tracking frameworks.

STARK~\cite{STARK} enhances template-search interaction by integrating transformer modules after CNN-based feature extraction. SwimTrack~\cite{SwinTrack} develops a SwinTransformer-based feature representation extractor and a motion-aware fusion module, explicitly incorporating motion cues during feature aggregation. ARTRack~\cite{ARTrack} adopts an encoder-decoder architecture to directly decode target location from visual features and coordinate tokens. ARTrackV2~\cite{ARTrackV2} extends this through joint trajectory-appearance autoregression, simultaneously predicting target positions and reconstructing target appearance to boost performance. ODTrack~\cite{ODTrack} leverages densely sampled video-clip and two temporal token propagation attention mechanisms to capture spatiotemporal information. HIPTrack~\cite{HIPTrack} maintains a memory bank updated via a FIFO strategy, enabling the model to utilize historical cues for more accurate tracking. SPMTrack~\cite{SPMTrack} proposes TMoE, introducing dynamic expert routing for adaptive relation modeling while enabling parameter-efficient fine-tuning. Despite their remarkable accuracy, these methods predominantly employ parameter-intensive architectures requiring GPU acceleration, limiting their applicability to lightweight trackers. To address these challenges, we design a video-level tracking framework that is more suitable for lightweight applications. Our model effectively captures spatiotemporal information with negligible sacrifice in inference speed.

\begin{figure*}[t]
	\centering
	\includegraphics[width=0.9\textwidth]{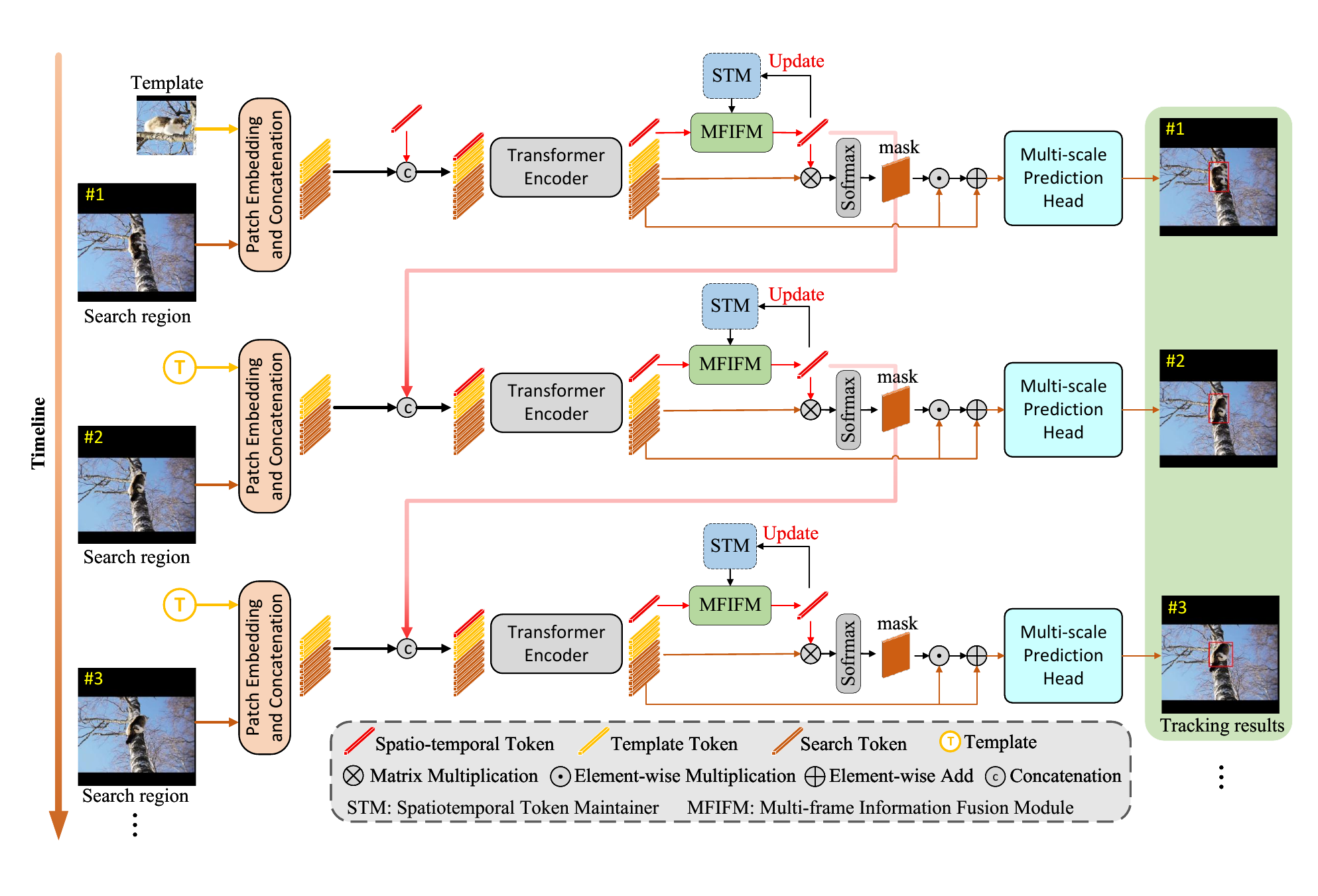} % Reduce the figure size so that it is slightly narrower than the column.
	\caption{The architecture of our STDTrack framework. It comprises four components: a transformer encoder for feature extraction, a Multi-frame Information Fusion Module (MFIFM) that enhances and refines spatiotemporal representations, a Spatiotemporal Token Maintainer (STM) for preserving high-quality temporal dependencies and an adaptive multi-scale prediction head.}
	\label{overall structure}
\end{figure*}

\textbf{Efficient Tracking Network.} Increasing research efforts have been directed toward developing lightweight tracking models capable of real-time operation on resource-constrained platforms. LightTrack~\cite{lighttrack} pioneers the application of neural architecture search (NAS) for object tracking, designing a lightweight search space and a dedicated search pipeline for tracking scenarios. FEAR~\cite{FEAR} proposes a lightweight tracker with a novel dual-template representation for object model adaptation. HCAT~\cite{HCAT} constructs a feature fusion network comprising a feature sparsification module and a hierarchical cross-attention transformer, which maintains competitive performance while reducing the computational amount. MixformerV2~\cite{MixFormerV2} employs knowledge distillation to yield a more unified and efficient tracker. FERMT~\cite{FERMT} introduces a non-deep feature extraction strategy within the backbone network and proposes a Dual Attention Unit (DAU) to address the performance degradation caused by model lightweighting. ORTrack~\cite{ORTrack} is designed for UAV tracking, integrating occlusion-robust representations and proposing an Adaptive Feature-Based Knowledge Distillation (AFKD) method to enhance accuracy and efficiency.

However, a performance gap persists between these lightweight trackers and high-performance counterparts. In this work, we focus on bridging this divide while maintaining real-time capability. To this end, we propose to exploit spatiotemporal features inherent in video sequences by incorporating reliable spatiotemporal dependencies during tracking, thereby significantly narrowing the accuracy gap while ensuring efficient computation.

\section{Approach}
We introduce STDTrack, a lightweight tracking framework that combines spatiotemporal dependencies, as shown in Figure \ref{overall structure}. In this section, we elaborate on the specific components of our model: transformer encoder, Multi-frame Information Fusion Module (MFIFM), Spatiotemporal Token Maintainer (STM) and the multi-scale prediction head.

\subsection{Overview}
The proposed STDTrack framework uses densely sampled video sequences, taking one template and video-clip as inputs. Unlike previous lightweight trackers that employ static training paradigms, our method adopts a dynamic training strategy where historical spatiotemporal tokens from preceding timesteps are preserved and propagated temporally to assist in current-frame localization. As depicted in Figure \ref{overall structure}, after feature extraction via the transformer encoder, spatiotemporal tokens are fed into the Multi-frame Information Fusion Module (MFIFM). This module leverages historically preserved dependencies in the Spatiotemporal Token Maintainer (STM) to refine and enhance feature representations. The augmented tokens are subsequently stored in the STM as compressed temporal context. Concurrently, we apply a mask-based search tokens enhancement mechanism guided by the spatiotemporal tokens, with the reinforced search tokens ultimately driving target localization predictions.

\subsection{Transformer Encoder}
The transformer encoder processes template tokens, search tokens, and spatiotemporal tokens as inputs. For each frame, we introduce a spatiotemporal token to summarize target-specific information. Specifically, template and search image undergo patch embedding and linear projection to generate template tokens and search tokens, denoted as $\textit{f}_z \in \mathbf{R}^{N_z\times D}$ and $\textit{f}_x \in \mathbf{R}^{N_x\times D}$ respectively. Here, $N_z = H_zW_z/P^2$, $N_x = H_xW_x/P^2$ and $P$ is the resolution of each patch. Learnable positional encoding are added to each token to provide relative positional information. Subsequently, spatiotemporal tokens, template tokens, and search tokens are concatenated and fed into the transformer encoder for representation learning. Note that for the first frame, we instantiate the spatiotemporal token as a learnable vector.

\begin{figure}[t]
	\centering
	\includegraphics[width=0.9\columnwidth]{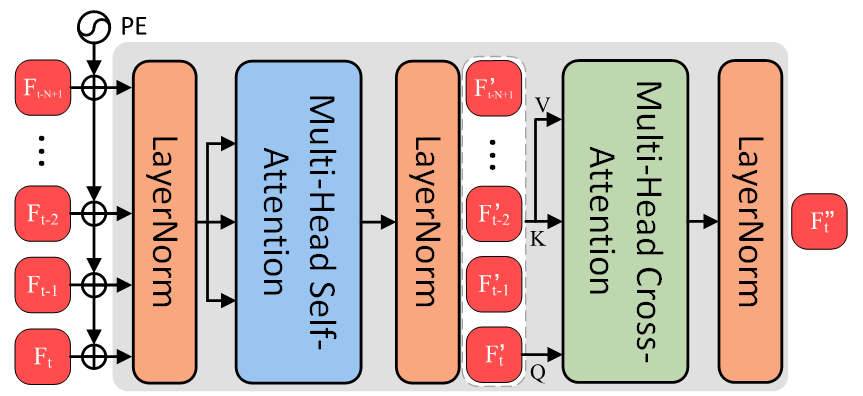} % Reduce the figure size so that it is slightly narrower than the column. Don't use precise values for figure width.This setup will avoid overfull boxes.
	\caption{Architecture of MFIFM. This module fuses the spatiotemporal feature vectors ${\left\{ \textbf{F}_t \right\}}_{t=1}^{N}$ (summarizing target information per frame) into an augmented representation $F''_t$ through temporal propagation of historical dependencies.}
	\label{multi-frame information fusion module}
\end{figure}

\subsection{Multi-frame Information Fusion Module}
During continuous tracking, historical target states capture appearance and morphological dynamics over time. By incorporating such historical context into current-frame tracking, the model gains enhanced perception of target variations, thereby improving localization accuracy. We therefore propose the Multi-frame Information Fusion Module (MFIFM), designed to endow the current spatiotemporal token with awareness of historical target states. Specifically, the MFIFM processes the spatiotemporal token $F_t$ from the transformer encoder alongside historically preserved feature vectors $\left\{ F_{t-1},...,F_{t-N+1} \right\}$ from the STM (detailed in next section). As illustrated in Figure \ref{multi-frame information fusion module}, fixed positional encoding~\cite{Transformer} are added to all tokens to preserve spatial ordering relationships. These tokens then undergo layer normalization before being processed by a multi-head self-attention layer. Leveraging the global modeling capability of attention mechanisms, each token perceives historical target states to obtain preliminarily enhanced tokens ${\left\{ \textbf{F'}_t \right\}}_{t=1}^{N}$. Subsequently, multi-head cross-attention facilitates interaction between the current token $F'_t$ and ${\left\{ \textbf{F'}_t \right\}}_{t=1}^{N}$, yielding the final enhanced spatiotemporal token $F''$. The process in the MFIFM can be described as:
\begin{equation}
	\begin{aligned}
		&F_{in} = LN({\left\{ \textbf{F}_t \right\}}_{t=1}^{N} + P_{fix})	\\
		&\left\{ \textbf{F'}_t \right\} = LN(MSA(Q=F_{in}, K=F_{in}, V=F_{in})) \\
		&F''_{t} = LN(MCA(Q=F'_t,K=\left\{ \textbf{F'}_t \right\},V=\left\{ \textbf{F'}_t \right\}))
	\end{aligned}
	\label{eq:2}
\end{equation}
where $P_{fix}$ represents the fixed positional encoding. To prevent this module from introducing excessive parameters that would compromise inference speed, we employ only a single self-attention layer and a single cross-attention layer.

\subsection{Mask-based Enhancement}
Before feeding the search token into the prediction head, we perform a mask-based feature enhancement operation to highlights the target regions in the search area while suppressing background interference, thereby enabling the prediction head to better perceive the target and generating more accurate tracking results (as validated in the ablation study).

As shown in Figure \ref{overall structure}, we generate the mask using MFIFM-enhanced spatiotemporal token and apply it to the search region features. Since the enhancement process incorporates spatiotemporal tokens, the prediction results are guided by spatiotemporal information, which also helps the model condense more reliable spatiotemporal dependencies during updates. A residual connection is employed to prevent information loss.

\subsection{Spatiotemporal Token Maintainer}
Quality assessment and filtering of spatiotemporal information are crucial during integration. Indiscriminately introducing dependencies would allow low-quality information to cause misjudgments of the target's current state, compromising tracking performance. We therefore devise a mechanism to harvest reliable spatiotemporal dependencies while dynamically pruning inferior features generated during tracking.

We construct a Spatiotemporal Token Maintainer (STM) of length $N$ to archive spatiotemporal tokens generated during tracking. At each timestep, the enhanced spatiotemporal token $F''_t$ from MFIFM and the score map produced by the prediction head are recorded in this maintainer. Since the final tracking result is determined by the maximum confidence value in the score map, the map directly reflects the quality of the current spatiotemporal token. We propose using target-background saliency as the criterion for assessing spatiotemporal feature quality. Specifically, after tracking the current frame, we compute the ratio of the maximum confidence value to the sum of all values in the score map as the quality $Q$:
\begin{equation}
	Q_t = \frac{max(score)}{\sum_{i=1}^{H}\sum_{j=1}^{W}score_{ij}}
	\label{eq:3}
\end{equation}
where the resolution of score map is $(H, W)$. A higher $Q$ indicates greater target saliency in the prediction map, signifying more reliable spatiotemporal dependencies. When the STM has not reaches its maximum capacity $N$, new tokens are stored directly. At full capacity, newly generated token is recorded while the maintainer simultaneously prunes the token with the lowest recorded $Q$ among existing entries. This update strategy ensures our model persistent access to the $N$ most reliable spatiotemporal dependencies during tracking.

\begin{figure}[t]
	\centering
	\includegraphics[width=0.9\columnwidth]{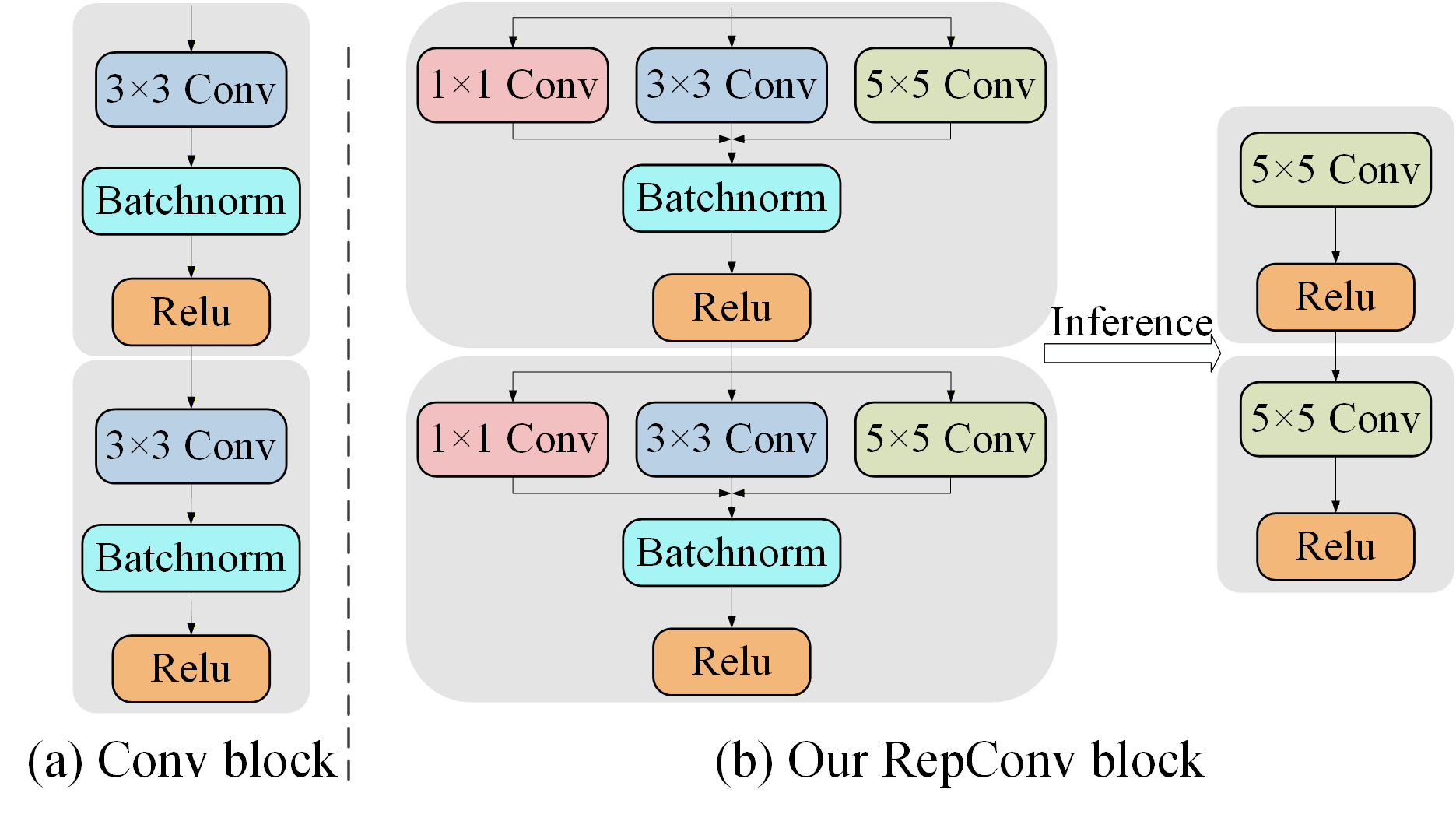} 
	\caption{RepConv block design. (a) Standard Conv block in the center head~\cite{OSTrack}. (b) Our proposed RepConv block employing structural re-parameterization technique. During inference, this technique merges multi-branch convolutional layers into a single convolution operation.}
	\label{multi-scale prediction head}
\end{figure}

\subsection{Multi-scale Prediction Head}
Contemporary trackers predominantly employ dual-branch (regression and classification) convolutional prediction heads~\cite{OSTrack,FERMT}. The classification branch predicts target center locations, while the regression branch estimates bounding-box offsets relative to the center. We observe that homogeneous convolutional kernel sizes in these heads restrict scale adaptability, causing suboptimal performance on targets of varying dimensions ($e.g.$, small objects). To address this limitation, we propose a multi-scale prediction head that captures multi-scale representations during training, enhancing robustness to scale variations. As illustrated in Figure \ref{multi-scale prediction head} (a), our head comprises stacked RepConv blocks. Within each block, we integrate additional $1 \times 1$ and $5 \times 5$ convolutional pathways to extract complementary features, subsequently fused through element-wise summation. The aggregated features finally used to generate a score map for classification branch, and local offsets and normalized bounding-box size for regression branch.

However, directly adding convolutional pathways would introduce excessive computational load, adversely affecting inference speed. We therefore utilize structural re-parameterization to merge multi-branch architectures into a single branch during inference. This mathematically equivalent transformation preserves model performance without compromise. For the proposed RepConv block, we first expand all convolution kernels to $5 \times 5$ dimensions through zero-padding to prepare for subsequent kernel fusion. Next, we consolidate the equivalent $5 \times 5$ convolution kernels and bias terms. Finally, we integrating convolution and batchnorm layer. This process is formally expressed as:
\begin{flalign}
	&\ \begin{aligned}
		&\widehat{K}_{:,:,i,j}^{(1)} = \left\{
		\begin{aligned}
			&K^{(1)} \quad &&i=2,j=2	\\
			&0 \quad &&otherwise	\\
		\end{aligned}
		\right.	\\
		&\widehat{K}_{:,:,i,j}^{(3)} = \left\{
		\begin{aligned}	
			&K_{:,:,i,j}^{(3)} \quad &&1\leq i\leq 3,1\leq j\leq 3	\\
			&0 \quad &&otherwise	\\
		\end{aligned}
		\right.	\\
		&\widehat{K}^{(5)} = K^{(5)}	\\
	\end{aligned}	&
	\label{eq:4}
\end{flalign} 
\begin{flalign}
	&\ \begin{aligned}
		&K_{merged} = \widehat{K}^{(1)} + \widehat{K}^{(3)} + \widehat{K}^{(5)}	\\
		&b_{merged} = \widehat{b}^{(1)} + \widehat{b}^{(3)} + \widehat{b}^{(5)}	\\
	\end{aligned}	&
	\label{eq:5}
\end{flalign}	
\begin{flalign}
	&\ \begin{aligned}
		&\widehat{K}_{merged} = \frac{\gamma}{\sqrt{\sigma^2 + \epsilon}}\cdot K_{merged}	\\
		&\widehat{b}_{merged} = \beta+\frac{\gamma\cdot(b_{merged}-\mu)}{\sqrt{\sigma^2 + \epsilon}}
	\end{aligned}	&
	\label{eq:6}
\end{flalign}
where $\widehat{K}$ denote the expanded convolution kernel. $\mu$, $\sigma^2$, $\gamma$, $\beta$ are the running mean, running variance, and learned scaling factor and bias of the BN layer. $\epsilon$ is a small positive constant.

\subsection{Training Loss}
The loss function combines focal loss~\cite{focalloss} for classification with L1 loss and GIoU loss~\cite{giou} for regression, formally expressed as:
\begin{equation}
	L_{total} = \lambda_{cls}L_{cls} + \lambda_{giou}L_{giou} + \lambda_{1}L_{1}
	\label{eq:7}
\end{equation}
where $\lambda_{cls}$, $\lambda_{giou}$ and $\lambda_{1}$ represent the weight values of each loss function.

\section{Experiments}
\subsection{Implementation Details}
Our tracker is implemented using Python 3.9.7 and PyTorch 2.0.0. All models were trained on a single NVIDIA GeForce RTX 4090.

\textbf{Training.} During training, we implement dense sampling of video sequences to provide richer spatiotemporal information. The maximum sampling interval is set to 200, with a video clip length of 8 frames. Video clips are reversed with 0.5 probability as a data augmentation strategy to enhance model robustness. The training dataset comprises GOT-10k~\cite{got}, TrackingNet~\cite{trackingnet}, LaSOT~\cite{lasot}, and COCO~\cite{coco}. We employ ViT-tiny~\cite{vit-tiny-pre} as our transformer encoder. Search images are resized to $256 \times 256$, while template is resized to $128 \times 128$ as model inputs. We optimize model parameters using AdamW with differential learning rate: 4e-5 for the transformer encoder and 4e-4 for the rest, coupled with 1e-4 weight decay. Training proceeds for 500 epochs, with learning rate decay initiating at epoch 400. For fair evaluation on GOT-10k, we adhere to the one-shot protocol, exclusively training on GOT-10k for 100 epochs and initiating decay at epoch 80.

\textbf{Inference.} During inference, the model first applies structural re-parameterization to transform the prediction head into the optimized architecture shown in Figure \ref{multi-scale prediction head} (b), enhancing computational efficiency. Spatiotemporal tokens are continuously extracted and stored in the STM. Upon reaching maximum capacity, the STM dynamically updates stored tokens based on quality assessments to ensure reliable dependencies. A Hanning window is applied to the predicted score map for motion trajectory smoothing. Our STDTrack achieves real-time performance at 41 FPS on CPU platforms and 192 FPS on GPU devices.

\begin{table*}[t]
	\renewcommand{\arraystretch}{1.1}
	\centering
	\setlength{\tabcolsep}{1mm}
	{\begin{tabular}{@{}c|c|ccc|ccc|ccc|cc@{}}
			\hline
			\multirow{2}{*}{Method} & \multirow{2}{*}{Source} & \multicolumn{3}{c|}{$\rm GOT{\text -}10k^*$} & \multicolumn{3}{c|}{TrackingNet} & \multicolumn{3}{c|}{LaSOT} & \multicolumn{2}{c}{Speed(fps)} \\
			\cline{3-13}
			&&$AO$ & $SR_{0.5}$ & $SR_{0.75}$ & $AUC$ & $P_{norm}$ & $P$ & $AUC$ & $P_{norm}$ & $P$ &GPU&CPU   \\
			\hline
			\multicolumn{13}{c}{CPU non-real-time Methods} \\
			\hline
			STARK-ST50&ICCV21&68.0&77.7&62.3&81.3&86.1&-&66.6&-&-&120&16	\\
			TransT&CVPR21&67.1&76.8&60.9&81.4&86.7&80.3&64.9&73.8&69.0&157&3	\\
			MixFormer-22k&CVPR22&70.7&80.0&67.8&83.1&88.1&81.6&69.2&78.7&74.7&106&8	\\
			$\rm OSTrack_{256}$&CVPR22&71.0&80.4&68.2&83.1&87.8&82.0&69.1&78.7&75.2&145&14\\
			$\rm ARTrack_{256}$&CVPR23&73.5&82.2&70.9&84.2&88.7&83.5&70.4&79.5&76.6&62&5\\
			$\rm ROMTrack_{256}$&ICCV23&72.9&82.9&70.2&83.6&88.4&82.7&69.3&78.8&75.6&99&10\\
			$\rm EVPTrack_{224}$&AAAI24&73.3&83.6&70.7&83.5&88.3&-&70.4&80.9&77.2&112&12\\
			$\rm LMTrack_{256}$&AAAI25&76.3&87.1&73.9&84.2&89.0&82.8&69.8&79.2&76.3&101&8\\
			SPMTrack-B&CVPR25&76.5&85.9&76.3&86.1&90.2&85.6&74.9&84.0&81.7&-&- \\
			\hline
			\multicolumn{13}{c}{CPU real-time Methods} \\
			\hline
			LightTrack&CVPR21&61.1&71.0&-&72.5&77.8&69.5&53.8&-&53.7&91&40\\
			STARK-Lighting&ICCV21&59.6&69.6&47.9&72.7&77.9&67.4&57.8&66.0&57.4&234&63\\
			FEAR-XS&ECCV22&61.9&72.2&-&-&-&-&53.5&-&54.5&252&85\\
			HCAT-B&ECCV22&65.1&76.5&56.7&76.6&82.6&72.9&59.3&68.7&61.0&371&21\\
			MixFormerV2-S&NIPS23&-&-&-&75.8&81.1&70.4&60.6&69.9&60.4&312&50\\
			SMAT&WACV24&64.5&74.7&57.8&78.6&84.2&75.6&61.7&71.1&64.6&168&36\\
			FERMT&ECCV24&\underline{69.6}&\underline{80.1}&\underline{63.2}&\underline{80.8}&\underline{85.9}&\underline{78.1}&\underline{65.1}&\underline{74.6}&\underline{69.1}&225&46\\
			S-Small&WACV25&64.6&75.1&-&78.4&83.5&74.6&60.7&-&62.2&254&45\\
			\hline
			STDTrack&ours&\textbf{71.3}&\textbf{81.6}&\textbf{66.2}&\textbf{81.4}&\textbf{86.1}&\textbf{79.0}&\textbf{67.3}&\textbf{77.3}&\textbf{72.2}&192&41\\
			\hline	
	\end{tabular}}
	\caption{Comparison on three test set of GOT-10k, Trackingnet, and LaSOT. $*$ denotes models that are trained exclusively on the GOT-10k dataset. The best two results are in \textbf{bold} and \underline{underline}, respectively.}
	\label{tab:1}
\end{table*}

\subsection{Comparison with the SOTA}
We evaluate our STDTRack on 6 challenging benchmarks: GOT-10k~\cite{got}, TrackingNet~\cite{trackingnet}, LaSOT~\cite{lasot}, AVisT~\cite{AVisT}, NFS\cite{nfs30}, and UAV123~\cite{uav}. To ensure fair comparison, we use the same evaluation metrics as~\cite{FERMT}, and all inference speed results are measured under identical hardware configurations.

\textbf{GOT-10k.} GOT-10k is a large-scale, generic object tracking benchmark containing over 10,000 videos with more than 1.5 million manually annotated bounding boxes. It features a diverse taxonomy of 563 object classes and emphasizes motion diversity to reflect real-world tracking challenges. As shown in Tab. \ref{tab:1}, STDTrack achieves state-of-the-art performance among real-time trackers, surpassing the previous best method FERMT by 1.7\% in AO. Notably, STDTrack outperforms the recent non-real-time tracker MixFormer while operating 1.8× faster on GPU and 5× faster on CPU platforms.

\textbf{TrackingNet.} TrackingNet offers an extensive dataset for large-scale training and evaluation of visual object trackers. It comprises over 30,000 videos with more than 14 million dense bounding box annotations. Its scale and diversity in object categories, scenes, and motion patterns make it suitable for assessing performance across varied real-world scenarios. As evidenced in Tab. \ref{tab:1}, our STDTrack surpasses the real-time tracker S-Small by significant margins, outperforming 2.0\%, 2.6\%, and 4.4\% in terms of AUC, normalized precision and precision. Simultaneously, it achieves competitive performance against non-real-time trackers ($e.g.$, STARK-ST50, TransT~\cite{TransT}), demonstrating exceptional generalization capability and robustness.

\begin{table}[t]
	\renewcommand{\arraystretch}{1}
	\centering
	\setlength{\tabcolsep}{1mm}
	{\begin{tabular}{@{}c|c|c|c|c@{}}
			\hline
			Method & Source & AVisT & NFS & UAV123  \\
			\hline
			LightTrack&CVPR21&40.4&56.5&61.7\\
			STARK-Lighting&ICCV21&-&59.6&62.0\\
			FEAR-XS&ECCV22&38.7&48.6&61.0\\
			HCAT-B&ECCV22&41.8&61.9&63.6\\
			MixFormerV2-S&NIPS23&39.6&61.0&63.4\\
			SMAT&WACV24&44.7&62.0&64.3\\
			FERMT&ECCV24&-&\underline{65.1}&67.5\\
			ORTrack&CVPR25&-&-&66.4\\
			S-Small&WACV25&\underline{47.9}&62.4&\underline{68.1}\\
			\hline
			STDTrack&ours&\textbf{53.9}&\textbf{65.3}&\textbf{68.4}\\
			\hline	
	\end{tabular}}
	\caption{Comparison of AUC metric on AVisT, NFS, and UAV123. The best two results are in \textbf{bold} and \underline{underline}, respectively.}
	\label{tab:2}
\end{table}

\textbf{LaSOT.} LaSOT is a high-quality benchmark specifically designed for long-term tracking evaluation. It contains 1400 sequences with an average length of over 2500 frames. As presented in Tab. \ref{tab:1}, our tracker establishes new SOTA among real-time methods. Compared to the latest method S-Small, we achieve significant improvements of 6.6\% in AUC and 10\% in precision (P), thereby validating the effectiveness of spatiotemporal feature integration within our framework.

\begin{table*}[tb]
	\renewcommand{\arraystretch}{1.1}
	\centering
	{\begin{tabular}{@{}c|c|c|c|c|ccc|c@{}}
			\hline
			\multirow{2}{*}{\#}&\multirow{2}{*}{Method}&\multirow{2}{*}{Params (M)}&\multirow{2}{*}{FLOPs (G)}&\multirow{2}{*}{FPS (CPU)}&\multicolumn{3}{c|}{GOT-10k}&\multirow{2}{*}{$\Delta (\%)$} \\
			\cline{6-8}
			&&&&&$AO(\%)$ & $SR_{0.5}(\%)$ & $SR_{0.75}(\%)$&  \\
			\hline
			1&Baseline &8.1&2.38&54& 69.2&79.6&64.8& - \\
			\hline
			2&+ spatiotemporal token&8.1&2.41&48&69.5&79.3&65.8& $+$ 1.0 \\ 
			\hline
			3&+ MFIFM&8.4&2.41&45& 70.2&80.4&65.7& $+$ 2.7 \\
			\hline
			4&+ mask-based enhancement &8.4&2.41&45& 70.7&80.7&\textbf{66.5}& $+$ 4.3 \\
			\hline
			5&+ multi-scale head&15.6&3.54&41& \textbf{71.3}&\textbf{81.6}&66.2& $+$ 5.5 \\
			\hline
	\end{tabular}}
	\caption{Ablation Study on GOT-10k. We incrementally integrating proposed modules into the baseline to quantify their contributions. $\Delta$ denotes the overall change of the three metrics against the baseline.}
	\label{ablation}
\end{table*}

\textbf{AVisT, NFS and UAV123.} AVisT is a highly challenging benchmark comprising 120 video sequences. It involving tracking object under extreme weather conditions such as heavy rain, heavy snow, and sandstorms. NFS is a benchmark designed for evaluating trackers under fast motion and motion blur challenges. This dataset includes 100 challenging videos with accurate bounding boxes, focusing on scenarios where rapid target movement is the primary difficulty. UAV123 is a benchmark captured from a low-altitude aerial perspective using UAVs, which contains 123 video sequences. These three datasets represent out-of-distribution (OOD) scenarios not encountered during training. As evidenced in Tab. \ref{tab:2}, our STDTrack achieves SOTA performance across all benchmarks. Notably, it attains a 6.0\% AUC improvement over the previous SOTA on AVisT, demonstrating exceptional generalization capability.

\subsection{Ablation Study}
\textbf{Importance of spatiotemporal token.} We first train a baseline model devoid of any proposed modules, which performs tracking using only an initial template and search image. We then introduce a spatiotemporal token to provide temporal context. The token generated in the current frame propagates to subsequent frames to guide feature extraction. From the first and second rows of Tab. \ref{ablation}, it can be observed that model performance improves after adding spatiotemporal token, validating the effectiveness of our token design.

\begin{table}[tb]
	\renewcommand{\arraystretch}{1.1}
	\centering
	\setlength{\tabcolsep}{1mm}
	{\begin{tabular}{@{}c|ccc@{}}
			\hline
			STM update mechanism&$AO(\%)$ & $SR_{0.5}(\%)$ & $SR_{0.75}(\%)$ \\
			\hline
			First-in first-out & 71.0&81.4&65.6 \\
			\hline
			Quality-based& \textbf{71.3}&\textbf{81.6}&\textbf{66.2} \\ 
			\hline
	\end{tabular}}
	\caption{Comparison of different STM update mechanism on GOT-10k.}
	\label{update mechanism}
\end{table}

\textbf{Study on MFIFM.} The Multi-frame Information Fusion Module (MFIFM) is one of the critical components in our model. It integrates the target's historical information into the current spatiotemporal dependencies, enabling the model to maintain a comprehensive understanding of the target's previous states during tracking. To further evaluate the effectiveness of this module, we conducted an ablation study to investigate its impact on model performance. As shown in row 2 and row 3 of Tab. \ref{ablation}, MFIFM achieves a substantial performance gain with only a minimal increase in parameters and computational cost. Specifically, it achieves gains of 0.7\% and 1.1\% on the AO and $\rm SR_{0.5}$ metrics, respectively. The results demonstrate that providing the model with richer historical information enhances its perception of the target's appearance variations throughout the tracking process, thereby contributing to more precise tracking.

\textbf{Effectiveness of mask-based enhancement.} To maximize the utility of the spatiotemporal tokens obtained from the current frame, we do not directly feed the output features from the transformer encoder into the prediction head. Instead, we first enhance these features using the spatiotemporal tokens, and then employ the enhanced features for target localization. As illustrated in Figure \ref{overall structure}, we generate a mask based on the interaction between the spatiotemporal token and the search tokens. This mask serves to highlight potential target regions and suppress background distractions. The effectiveness and feasibility of this operation are validated by the results in rows 3 and 4 of Tab. \ref{ablation}. Specifically, incorporating our proposed mask-based enhancement further boosts model performance, yielding an additional improvement of 0.5\% in the AO metric.

\begin{table}[tb]
	\renewcommand{\arraystretch}{1.1}
	\centering
	{\begin{tabular}{@{}c|c|ccc@{}}
			\hline
			\textbf{\#}&capacity&AO$(\%)$&$SR_{0.5}$$(\%)$&$SR_{0.75}$$(\%)$ \\
			\hline
			$\quad1\quad$&2&69.6&79.3&65.4 \\
			$\quad2\quad$&4&70.9&81.2&65.9 \\ 
			$\quad3\quad$&6&\textbf{71.3}&\textbf{81.6}&\textbf{66.2} \\ 
			$\quad4\quad$&8&69.9&79.9&65.2 \\
			\hline
	\end{tabular}}
	\caption{Comparison of different STM capacities on GOT-10k.}
	\label{ablation on capacity}
\end{table}

\textbf{Multi-scale prediction head.} The multi-scale prediction head enhances the model's robustness by adapting to targets of varying sizes through receptive fields at different scales. To investigate its contribution, we design experiment to assess its impact on performance. As shown in the forth and fifth rows of Tab. \ref{ablation}, employing the multi-scale prediction head leads to a improvement in overall performance. Despite the additional learnable parameters introduced by this module, the use of structural re-parameterization limits the inference speed penalty to a minor cost. These experimental results demonstrate its positive impact on the overall performance of the tracker.

\textbf{The update mechanism of STM.} To investigate the impact of different approaches to maintaining spatiotemporal dependencies, we designed two distinct STM update mechanisms. As presented in Tab. \ref{update mechanism}, we compare a First-In-First-Out (FIFO) strategy with our proposed quality-based update mechanism. Experimental results indicate that the quality-based approach achieves superior performance across all three evaluation metrics on GOT-10k. While the FIFO mechanism ensures the model consistently accesses the most recent information, its lack of discriminative filtering allows low-quality spatiotemporal dependencies to mislead the model, ultimately degrading performance. In contrast, our proposed quality-based update mechanism reliably provides the model with high-fidelity spatiotemporal dependencies, effectively enhancing tracker robustness.

\textbf{The capacity of STM.} We design a Spatiotemporal Token Maintainer (STM) to store the target's historical state representations across past frames. It is hypothesized that larger STM capacity provides more comprehensive target context for the tracker, potentially enhancing performance. However, tracking inaccuracies in individual frames introduce discrepancies between stored states and groundtruth. These errors accumulate temporally within the maintainer. Excessively large capacity heightens the risk of storing corrupted records. We therefore conduct an ablation study on STM capacity to identify an optimum range.As shown in Tab. \ref{ablation on capacity}, the experimental results align with our hypothesis: tracker performance improves progressively with increasing STM capacity, but deteriorates rapidly beyond a certain threshold. For this work, we select a STM capacity of 6.

\begin{figure}[t]
	\centering
	\includegraphics[width=1\linewidth]{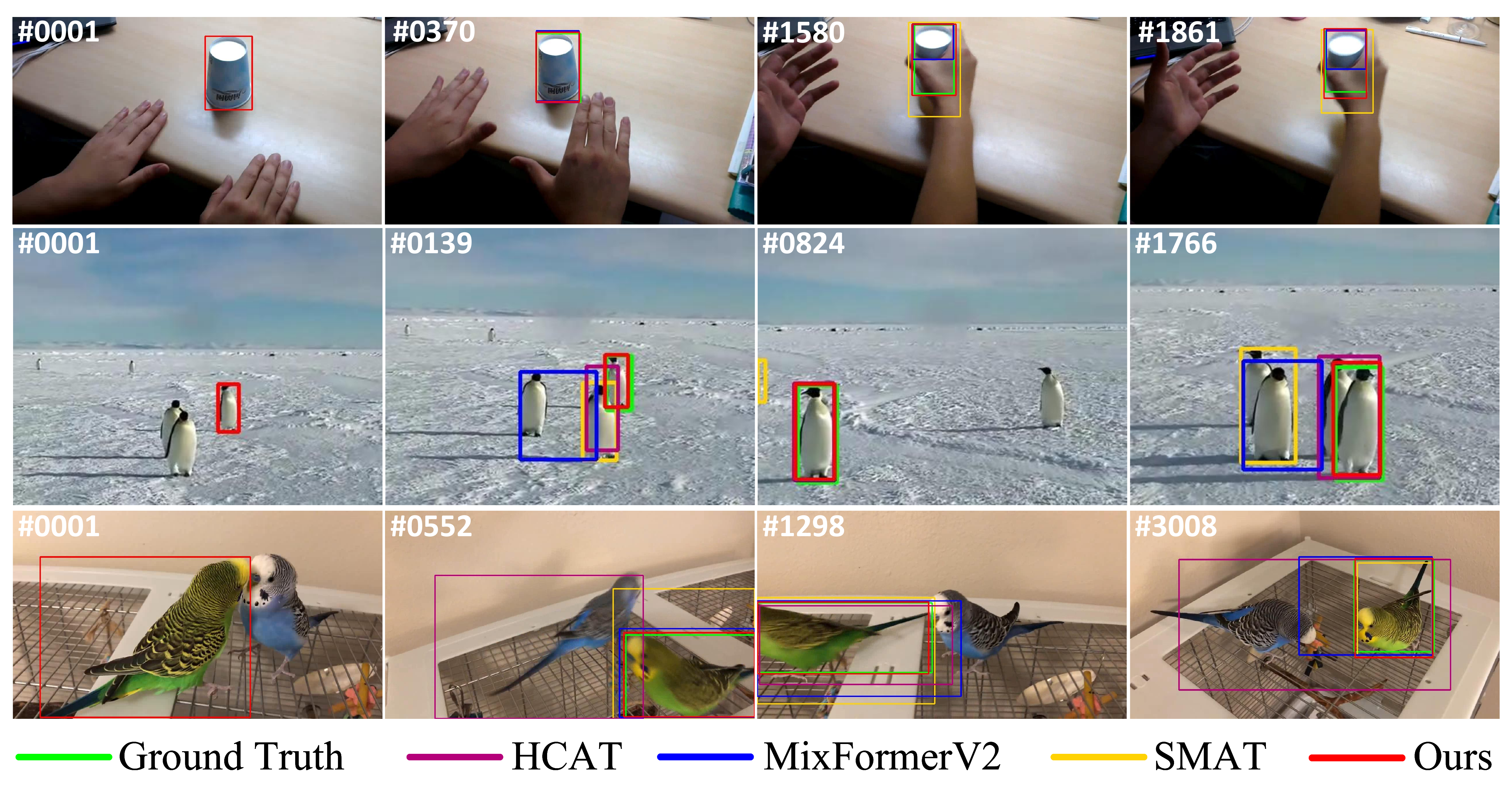}
	\caption{Qualitative comparison of our STDTrack and other SOTA trackers on LaSOT benchmark.}
	\label{Visualizations}
	% \vskip -0.1in
\end{figure}

\subsection{Visualization}
\textbf{Visualization.} To demonstrate the effectiveness of our STDTrack, we compare it against three state-of-the-art lightweight trackers on the LaSOT benchmark. As shown in Figure \ref{Visualizations}, our approach achieves superior performance. This advantage stems from our integration of reliable spatiotemporal dependencies, which provide historical target states during challenging scenarios ($e.g.$, distractor interference), thereby enhancing model robustness.

\begin{figure}[t]
	\centering
	\includegraphics[width=1\linewidth]{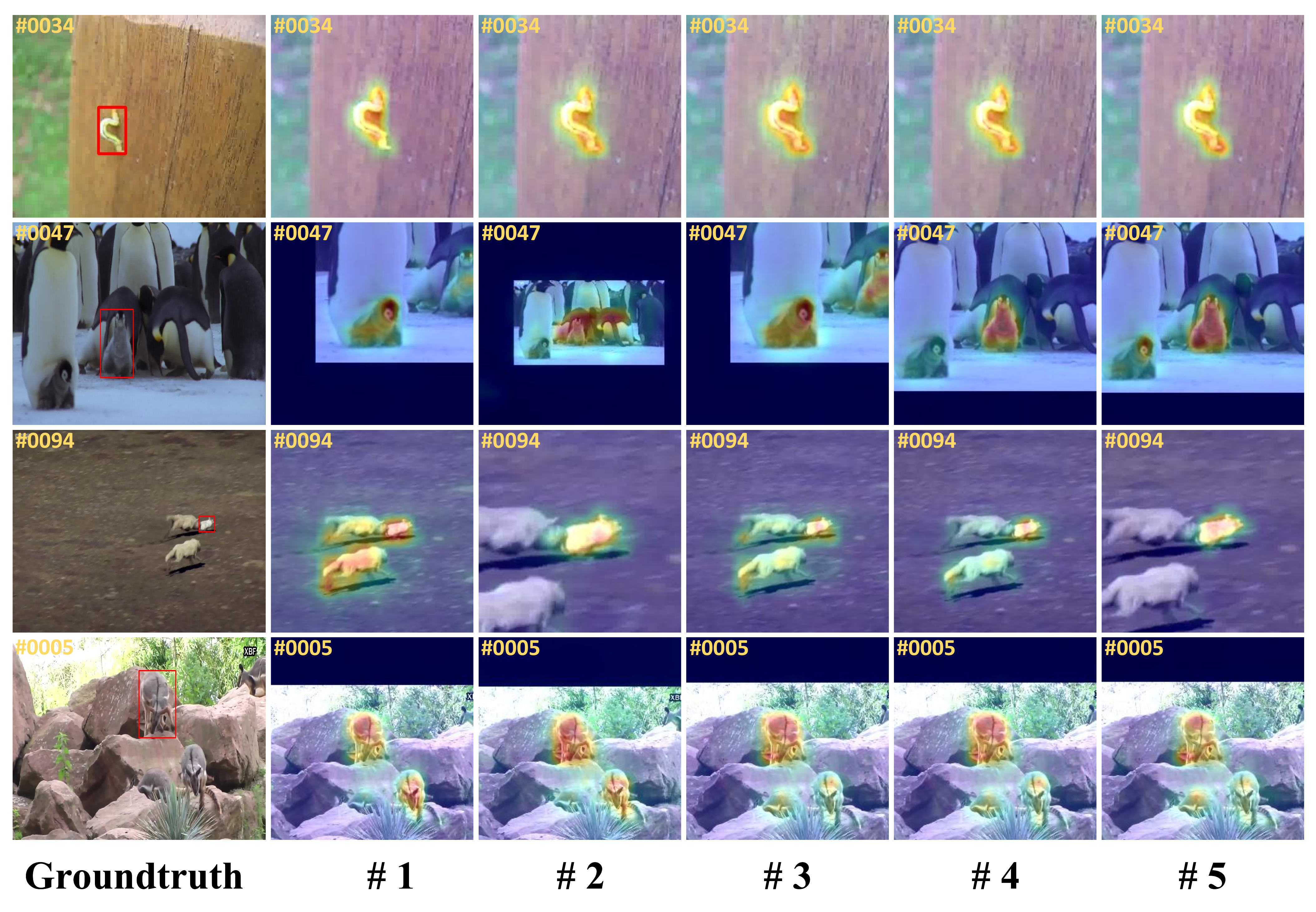} % Reduce the figure size so that it is slightly narrower than the column.
	\caption{A comparison of heatmaps between the baseline and models with the proposed modules integrated progressively. Each column corresponds to a model in Tab. \ref{ablation}.}
	\label{heatmap}
\end{figure}

Furthermore, we incrementally integrate the proposed modules into the baseline model and compare the differences in attention heatmaps generated during tracking. As shown in Figure \ref{heatmap}, the first column marks the groundtruth position of target, while each subsequent column corresponds to the heatmap generated by the respective model listed in Tab. \ref{ablation}. It can be observed that after incorporating our proposed modules, the models exhibit enhanced resistance to environmental distractions compared to the baseline, allowing them to focus more accurately on the target and achieve improved tracking performance. Specifically, when the target is occluded or disturbed by similar objects, the baseline model inevitably allocates attention to surrounding distractors. However, as the proposed modules are progressively added, the model gains the ability to perceive spatiotemporal information, leading to a gradual convergence of attention, which ultimately focuses entirely on the target. This observation demonstrates the effectiveness of our proposed modules and highlights the robustness and resistance to distractors of the overall approach.

%\textbf{Limitation.} This work introduces reliable spatiotemporal dependencies into lightweight trackers to enhance performance. Despite significant improvements, both the spatiotemporal modeling components and multi-scale prediction head inevitably increase model parameters. While structural re-parameterization mitigates computational burden during inference, some speed degradation persists. A promising future direction involves knowledge distillation: training a teacher tracker with robust spatiotemporal modeling capabilities, then transferring its knowledge to a lightweight student model. This approach could maintain competitive accuracy while substantially boosting inference speed.

\section{Discussion and Conclusion} 
In this work, we present STDTrack, the first video-level framework designed for lightweight tracking. We densely sample video sequences and propagate spatiotemporal features during training, providing rich contextual information to the model. Specifically, at each timestep, we introduce a spatiotemporal token that propagates temporally to summarize target-specific information. Furthermore, we propose MFIFM to extend temporal perception to historical frames while adaptively enhancing current features. The constructed STM stores frame-level spatiotemporal tokens and employs quality-based updates to ensure dependency reliability. Additionally, our multi-scale prediction head adapts to targets of varying dimensions. These innovations collectively enable STDTrack to achieve substantial improvements while maintaining real-time capability. Extensive experiments demonstrate state-of-the-art results across six benchmarks. We anticipate this work will inspire future research in efficient visual tracking.

\textbf{Limitations and future works.} Our tracker may fail in certain extreme scenarios, such as when the target moves completely outside the search region or under drastic scene changes. To address this, we plan to incorporate self-assessment and global search mechanisms to enable recovery from tracking failures. Additionally, the current model does not employ compression methods like distillation or pruning. Judicious application of these techniques could make our tracker faster and more lightweight. These directions represent important avenues for our future research.
\clearpage 

\bibliography{aaai2026}

\end{document}